\documentclass[letterpaper]{article} 
\usepackage{aaai24}  
\usepackage{times}  
\usepackage{helvet}  
\usepackage{courier}  
\usepackage[hyphens]{url}  
\usepackage{graphicx} 
\usepackage{natbib}  
\usepackage{caption} 
\usepackage{algorithm}
\usepackage{algorithmic}

\usepackage{multirow}
\usepackage{enumerate}
\usepackage{amssymb}
\usepackage{subfig}
\usepackage{amsmath,bm}
\usepackage{booktabs}
\usepackage{color}

\usepackage[capitalize]{cleveref}
\crefname{section}{Sec.}{Secs.}
\Crefname{section}{Section}{Sections}
\Crefname{table}{Table}{Tables}
\crefname{table}{Tab.}{Tabs.}

\usepackage{newfloat}
\usepackage{listings}
\DeclareCaptionStyle{ruled}{labelfont=normalfont,labelsep=colon,strut=off} 
\floatstyle{ruled}
\newfloat{listing}{tb}{lst}{}
\floatname{listing}{Listing}
\title{No Regularization is Needed: An Efficient and Effective Model for Incomplete Label Distribution Learning}
\author{
    Xiang Li\textsuperscript{\rm 1}, Songcan Chen\textsuperscript{\rm 1*}
}
\affiliations{
    \textsuperscript{\rm 1} College of Computer
    Science and Technology/College of Artificial Intelligence\\
    Nanjing University of Aeronautics and Astronautics\\

}

\usepackage{bibentry}

\begin{document}

\maketitle

\begin{abstract}
Label Distribution Learning (LDL) assigns soft labels, a.k.a. degrees, to a sample. In reality, it is always laborious to obtain complete degrees, giving birth to the Incomplete LDL (InLDL). However, InLDL often suffers from performance degeneration. To remedy it, existing methods need one or more explicit regularizations, leading to burdensome parameter tuning and extra computation. We argue that label distribution itself may provide useful prior, when used appropriately, the InLDL problem can be solved without any explicit regularization. In this paper, we offer a rational alternative to use such a prior. Our intuition is that large degrees are likely to get more concern, the small ones are easily overlooked, whereas the missing degrees are completely neglected in InLDL. To learn an accurate label distribution, it is crucial not to ignore the small observed degrees but to give them properly large weights, while gradually increasing the weights of the missing degrees. To this end, we first define a weighted empirical risk and derive upper bounds between the expected risk and the weighted empirical risk, which reveals in principle that weighting plays an implicit regularization role. Then, by using the prior of degrees, we design a weighted scheme and verify its effectiveness. To sum up, our model has four advantages, it is 1) \textit{model selection free}, as no explicit regularization is imposed; 2) with closed form solution (sub-problem) and easy-to-implement (a few lines of codes); 3) with \textit{linear} computational complexity in the number of samples, thus scalable to large datasets; 4) competitive with state-of-the-arts even without any explicit regularization.
\end{abstract}

\section{Introduction} \label{sec_intro}
In real applications, labels may be associated with a sample to some degree, thus soft labels are preferred rather than the hard ones to describe the label ambiguity \cite{rupprecht2017learning}, where Label Distribution Learning (LDL) \cite{geng2016label} originated from. LDL is a learning paradigm that assigns a sample with different label description degrees, \textit{i.e.}, the probabilities of a sample belonging to different labels, which satisfy the probability simplex constraint. To date, LDL has successful and wide applications in facial age estimation \cite{shen2017label, wen2020adaptive, zhang2021practical}, facial expression recognition \cite{jia2019facial,chen2020label,zhao2021robust}, multi-label classification \cite{zhang2019leveraging, xu2020partial, lu2022predicting}. Despite its success, obtaining complete label degrees is always laborious and challenging in real-world, thus the desire to get rid of such predicament drives the emergence of Incomplete LDL (InLDL). 

Under the setting of InLDL, only part of label description degrees need to be given, which indeed reduces the costs of assigning degrees. However, such incomplete label degrees also make the supervised information insufficient, thus will inevitably degenerate the performance. To assuage this issue, existing InLDL methods always have to make various assumptions, which are then translated into one or more explicit regularization terms in their modeling to relieve the incompleteness. For example, in \cite{xu2017incomplete}, the authors assume that the matrix formed by the label degrees is low-rank to characterize the correlation between labels, and adopt the trace norm as a regularization term to cater for the low-rankness; in \cite{wang2023label}, the authors assume that the predictions of their model lie in the same manifold, and exploit both the global and local label correlations with three different regularization terms. Obviously, each imposed regularization term associates with a hyper-parameter, which needs tuning and extra computation cost for model selection. Besides, for the convenience of optimization, some assumptions could only be approximated by the regularization terms, for example, minimizing the trace norm is an approximation for pursuing the low-rankness \cite{mishra2013low}, and sometimes such an approximation leads to suboptimal results \cite{dai2014rank}.

\begin{figure} [t]
	\centering
	\subfloat [Orignal image.] {\label{fig1a} \includegraphics[width=0.43\linewidth,height = 2.7cm] {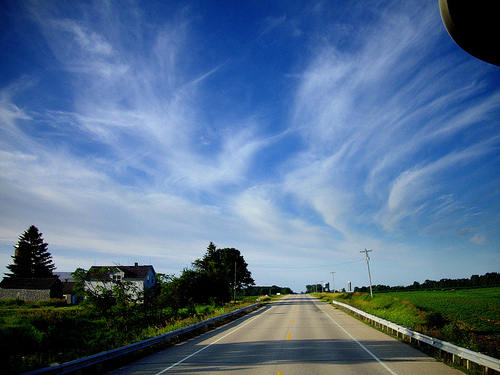}} \quad
	\subfloat [Incomplete label distribution.] {\label{fig1b} \includegraphics[width=0.43\linewidth, height = 2.7cm] {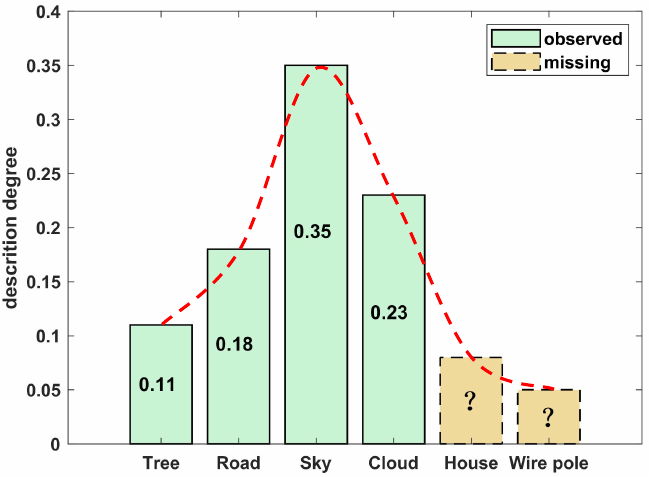}}
	\caption{An illustration of incomplete label distribution learning. Figure 1(a) is an image from real-world and Figure 1(b) is the corresponding label distribution. Note that, the degrees of the labels ``House'' and ``Wire pole'' are missing.} 
	\label{fig1}
\end{figure}

Different from existing InLDL methods that focus on learning the correlations between labels, we argue that label distribution itself may provide useful even sufficient prior knowledge. Concretely, it not only contains the correlations between labels but also reflects other relationships, such as the ranking orders of label degrees in a sample. Focusing only on the label correlations while ignoring other relationships is a waste of such a useful prior. Thus, this motivates us to put more emphasis on the prior itself, more fascinatingly, we find that when used appropriately, the InLDL problem can be solved even without any explicit regularization. In this paper, we offer an alternative to make rational use of the prior. Our starting point is from such an intuition that labels with large degrees are likely to receive more concern, whereas those with small degrees are easily overwhelmed by the large ones or even overlooked. As shown in \Cref{fig1a}, when looking at such a picture, labels with large degrees, such as ``Sky'' and ``Cloud'', are more easily to be noticed, whereas the label with small degree, such as ``Tree'' is easily overlooked, since the attention of one person is mostly drawn by the ``Sky'' and ``Cloud''. In the scenario of InLDL, labels with missing degrees are completely neglected. As shown in \Cref{fig1b}, the degrees of ``House'' and ``Wire pole'' are missing, thus their supervised information is heavily insufficient. From Figure 2, it is evident that the mean relative errors are higher for degrees less than 0.5 compared to those degrees greater than 0.5. This observation indicates that small degrees are underfitted in comparison to those large degrees. Note that in label distribution learning, the goal is to learn a distribution of all labels rather than a distribution of those labels with large degrees. Therefore, labels with small degrees, especially the missing ones, should be paid more attention so as to learn a more accurate label distribution. The question, then, is how to put more emphasis on those small and missing degrees? A natural and straightforward solution is to design a weighting scheme that incorporates the degree prior, since the degree itself not only can be used as weights inherently, but also reflects the relationship between labels.

\begin{figure} [t]
	\centering
	\subfloat [Flickr.] {\label{fig2a} \includegraphics[width=0.43\linewidth,height = 2.7cm] {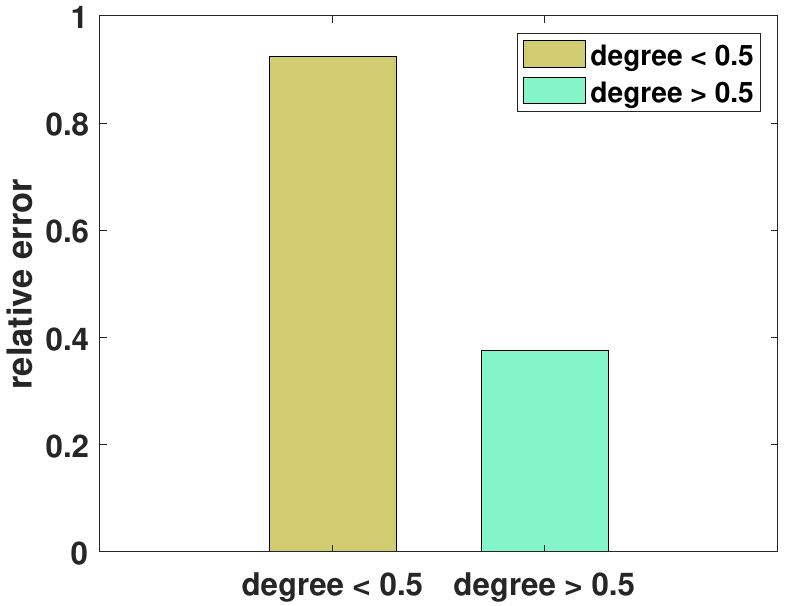}} \quad
	\subfloat [Fbp5500.] {\label{fig2b} \includegraphics[width=0.43\linewidth, height = 2.7cm] {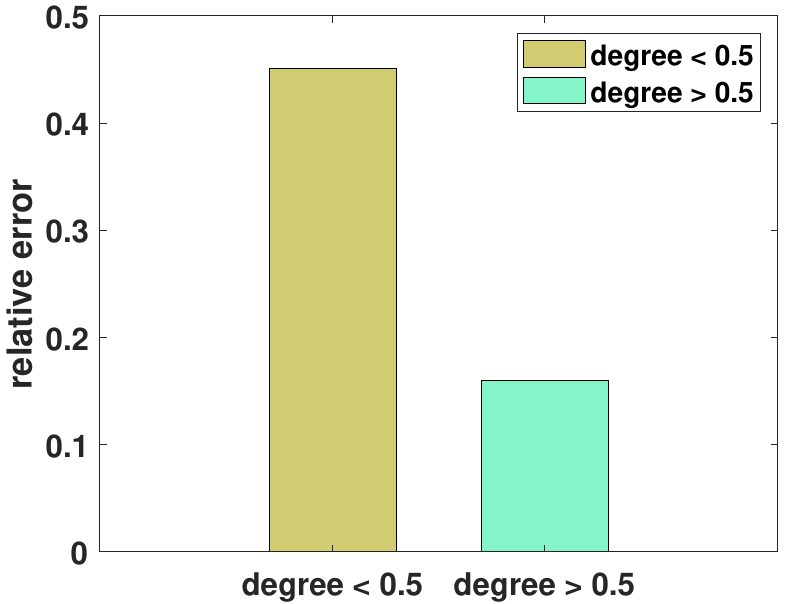}}
	\caption{The mean relative errors are computed for degrees less than 0.5 and degrees greater than 0.5 on two datasets. Figure 2(a) presents the results for the Flickr dataset, and Figure 2(b) displays the results for the Fbp5500 dataset.} 
	\label{fig2new}
\end{figure}
Inspired by the above intuition and with the aim to learn an accurate label distribution for all labels, it is important not to ignore the small observed degrees but to impose large weights on them to avoid being overwhelmed by the large degrees. Moreover, the weights of the missing degrees should be gradually increased, because as the training progresses, these degrees should be more and more reliable. Before designing a specific weighting scheme, we first define a weighted empirical risk and make some theoretical analyses, and then conduct an empirical case study to verify the effectiveness of our proposed model. Specifically, we provide theoretical guarantees by deriving data-dependent upper bounds between the expected risk and the weighted empirical risk. It is worth noting that the upper bounds \textit{explicitly} depend on the $\ell_\infty$ norm of the weighted matrix, implying that such weighting plays a role of \textit{implicit} regularization as no explicit regularization is imposed. After that, we naturally design a weighted matrix by directly using the given degree prior as weights on the losses such that small degrees are attached with large weights. Note that our weighted matrix is not from optimization, but directly from the degree prior. As a result, we can save additional overhead of learning a weighted matrix. Subsequently, we utilize the Alternating Direction Method of Multipliers (ADMM) \cite{boyd2011distributed} to optimize our model. By that, we derive the closed form solution of each sub-problem and the codes can be easily implemented in just a few lines. Interestingly, the computational complexity is linear in the number of samples, making our model fast and scalable to large datasets. 

To sum up, our main contributions are threefold:
\begin{enumerate}[(1)]
	\item  We propose an efficient, effective, and easy-to-implement Weighted model for InLDL, abbreviated as WInLDL, which is free of any explicit regularization, to the best of our knowledge, this is the first time in the field of InLDL.
	\item We theoretically derive a data-dependent upper bound between the expected risk and the weighted empirical risk with the help of Rademacher complexity, which contains the classical risk bounds under single-label and unweighted settings as our special cases.
	\item We empirically verify the effectiveness of WInLDL on ten real-world datasets, and experiments show that even without any explicit regularization, it is competitive with state-of-the-art methods.  
\end{enumerate}

\section{Related work}
In this section, we review the most relevant works to ours. InLDL was first proposed by Xu and Zhou \cite{xu2017incomplete} to address the problem with incomplete annotations. They assume that the matrix formed by the label degrees is low-rank to combat the incompleteness, and adopt the trace norm as a regularization term to formulate the label correlations. Later, in \cite{jia2019facial}, the authors assume the clusters of samples are low-rank and also utilize the trace norm to characterize such local label correlations. Recently, in \cite{wang2023label}, the authors argue that the low-rank assumption may not hold, instead, they assume that the predictions of their model lie on the same manifold whose structure may encode the correlations among labels. Further, they exploit both the global and local correlations to learn the label distribution in the InLDL setting. All the above methods focus on mining label correlations while ignore the fact that label distribution itself provides useful even sufficient prior knowledge, as we dissected in the Introduction. We contend that the degree prior should not be overlooked but rather utilized rationally. To the best of our knowledge, \cite{wang2021re} is the only work that considers the label degrees to do the weighting. Compared with our work, there are three main differences, 1) their task focuses on classification, the goal is to learn the top label(s), thus they put large weights on large degrees, which will be verified in \Cref{tab6} that in the InLDL setting, such a weighting performs worse than our WInLDL; 2) they discard the weighting scheme in their theoretical analysis, thus they do not provide a theoretical guarantee for weighting, which is a main contribution of this paper; 3) they use the product between the entropy ($E_{\bf{x}}=- \textstyle {\sum_{y \in \mathcal{Y}} d_{\bf{x}}^{y} \ln d_{\bf{x}}^{y}}$) of degree and degree itself as the weights. While in the InLDL setting, the missing degrees are set to 0, and $\ln d_{\bf{x}}^{y}$ will be meaningless in such a situation. Therefore, their weighting scheme cannot be applied to the InLDL setting. There are other works in the field of InLDL \cite{jia2021semi, teng2021incomplete, zhang2022safe, qian2022incomplete, qian2022local} that either adopt different settings or focus on different tasks, such as feature selection. Consequently, they are not highly relevant to the scope of this work. Due to the page limitations, we have omitted discussing these works here, and interested readers are referred to these literatures for further details.

\section{Proposed method}
\subsection{Problem setting}
Let $\mathcal{X} \subseteq \mathbb{R}^{k}$ be the feature space and $\mathcal{Y}=\{y_{1},y_{2},\cdots, y_{C}\}$ be the label space, where $k$ is the dimension of the feature, and $C$ is the number of labels. In LDL, each sample $\mathbf{x} \in \mathcal{X}$ is assigned with a label distribution $\mathbf{d_x} = [{d}_\mathbf{x}^{y_1},{d}_\mathbf{x}^{y_2},\cdots,{d}_\mathbf{x}^{y_C}]^\top$, where ${d}_\mathbf{x}^{y_i}$ is called the label description degree, which indicates the probability of sample $\mathbf{x}$ belonging to the $i$-th label. Note that, $\mathbf{d_x}$ satisfies the probability simplex constraints, \textit{i.e.}, ${d}_\mathbf{x}^{y_j}\ge 0$ and ${\textstyle \sum_{j=1}^{C}}{d}_\mathbf{x}^{y_j}=1$. Given a training dataset ${S} = \{ (\mathbf{x}_i,\mathbf{d}_{\mathbf{x}_i}) \}_{i=1}^{N}$, where $N$ is the number of samples, $\mathbf{X} = [\mathbf{x}_1,\mathbf{x}_2,\cdots,\mathbf{x}_N]^\top \in \mathbb{R}^{N \times k} $ is the feature matrix, $\mathbf{D} = [\mathbf{d}_{\mathbf{x}_1},\mathbf{d}_{\mathbf{x}_2},\cdots,\mathbf{d}_{\mathbf{x}_N}]^\top \in \mathbb{R}^{N \times C} $ is the label distribution matrix. The goal of LDL is to learn a function $f$ : $\mathcal{X} \mapsto \mathbb{R}^C$ , which minimizes the difference between the prediction of $f$ and the ground-truth label distribution, \textit{i.e.}, $\textstyle \min_{f}{\mathcal{L} (f(\mathbf{X}),\mathbf{D})} $, where $\mathcal{L}$ is the loss function.

In the scenario of InLDL, label degrees may be incomplete, following the setting of \cite{xu2017incomplete}, we also assume that entries in label distribution matrix $\mathbf{D}$ are uniformly random missing. The reason for following the uniformly missing assumption instead of assuming that small degrees are more likely to be missing is to eliminate any potential bias that would favor our approach, which ensures a fair and unbiased comparison. Let $\Omega \in [N] \times [C]$ and $U \in [N] \times [C]$ denote the indices of the observed and the unobserved entries sampled uniformly from $\mathbf{D}$, respectively. The unobserved entries of the label distribution matrix are set to 0, \textit{i.e.}, the observed label matrix $\widetilde{\mathbf{D}}$ can be defined as, $\forall (i,j) \in [N] \times [C]$,

\begin{align*}
\widetilde{\mathbf{D}}_{ij} =
\begin{cases}
\mathbf{D}_{ij}  & \text{if  } (i,j) \in \Omega\\
0  & \text{if  } (i,j) \in U.
\end{cases}
\end{align*}
Then in InLDL, the goal is to $\textstyle \min_{f}{\mathcal{L} (f(\mathbf{X}),\widetilde{\mathbf{D}})}$.

\subsection{Weighted empirical risk}
From the analysis in Introduction, to learn an accurate label distribution for all labels, we need to emphasize more on the small degrees. A natural and straightforward approach is assigning large weights on the losses of those small degrees, which results in an objective for minimizing the weighted empirical risk. Formally, the definition is detailed below.

\textbf{Definition 1.} Given a training dataset ${S} = \{ (\mathbf{x}_i,\mathbf{d}_{\mathbf{x}_i}) \}_{i=1}^{N}$, a class of functions $\mathcal{F}$, a loss function $\mathcal{L}$, and a weighted matrix $\mathbf{P} \in \mathbb{R}^{N \times C} $, the weighted empirical risk of function $f \in \mathcal{F}$ can be defined as:
\begin{align*}
\hat{\mathcal{R}}_{S}(f) = \frac{1}{N} \sum_{i=1}^{N} \sum_{j=1}^{C} \mathbf{P}_{ij} \mathcal{L}(f(\mathbf{x}_i)_j,\mathbf{D}_{ij}), 
\end{align*}
where $f(\mathbf{x}_i)_j$ denotes the $j$-th element of $f(\mathbf{x}_i)$. 
Suppose the population data follow an underlying probability distribution $\mathcal{D}$, then the expected risk can be written as:
\begin{align*}
\mathcal{R}_{\mathcal{D}}(f) = \mathbb{E}_{S \sim \mathcal{D}} [\hat{\mathcal{R}}_{S}(f)] 
\end{align*}
Before deriving our risk bound, we also provide the definition of empirical Rademacher complexity and Rademacher complexity for self-containment.

\textbf{Definition 2.} \cite{koltchinskii2001rademacher,bartlett2002rademacher} Let $\mathcal{F}$ be a class of functions, ${S} = \{ (\mathbf{x}_i,\mathbf{d}_{\mathbf{x}_i}) \}_{i=1}^{N}$ be a fixed size dataset with $N$ samples, and $\mathcal{L}$ be a loss function. Then, the empirical Rademacher complexity of $\mathcal{F}$ with respect to the sample set ${S}$ is defined as:
\begin{align*}
\hat{\mathfrak{R}}_{{S}}(\mathcal{F}) = \mathop{\mathbb{E}}\limits _{\bm{\sigma}}[\sup \limits _{f \in \mathcal{F}} \frac{1}{N}\sum_{i=1}^{N} {\sigma}_i \mathcal{L}(f(\mathbf{x}_i),\mathbf{d}_{\mathbf{x}_i})],
\end{align*}
where $\bm{\sigma} = ({\sigma}_1, {\sigma}_2, \cdots, {\sigma}_N)^\top$, with independent rademacher random variables ${\sigma}_i$s uniformly taking values in $\{-1,+1\}$. Then the Rademacher complexity
of $\mathcal{F}$ is the expectation of the empirical Rademacher complexity over all samples of size $N$ drawn according to the distribution $\mathcal{D}$:
\begin{align*}
{\mathfrak{R}}_{N}(\mathcal{F}) = \mathop{\mathbb{E}}\limits _{S\sim \mathcal{D}^N} [\hat{\mathfrak{R}}_{\mathcal{S}}(\mathcal{F})].
\end{align*}
In the following, we will derive an upper bound between the expected risk and the weighted empirical risk with the help of Rademacher complexity.

\textbf{Theorem 1.} Let $\mathcal{F}$ be a class of functions, $\mathcal{L} = {\textstyle \sum_{j=1}^{C}\ell(f(\mathbf{x})_j,d_\mathbf{x}^{y_j})}$ is a loss function, where $\ell$ is bounded by a constant $B$, and $\left \|  \mathbf{P}\right \|_\infty = \max_{ij} \mathbf{P}_{ij}$. Then, $\forall \delta > 0$, with probability at least $1 - \delta$ over the draw of an \textit{i.i.d.} sample $S$ of size $N$ from the distribution $\mathcal{D}$, the following bound holds for all $f \in \mathcal{F}$:
\begin{align} \label{eq1}
\mathcal{R}_{\mathcal{D}}(f) \le \hat{\mathcal{R}}_{S}(f) + 2\left \|  \mathbf{P}\right \|_\infty {\mathfrak{R}}_{N}(\mathcal{F}) + CB\left \|  \mathbf{P}\right \|_\infty\sqrt{\frac{log\frac{1}{\delta} }{2N} }.
\end{align}
The proof of this theorem primarily relies on the McDiarmid inequality \cite{mcdiarmid1989method}, and details can be found in the Appendix. Note that, the above bound can be seen as a generalization of the single-label and unweighted cases. When $C=1$ (single-label), and $\mathbf{P}$ is a matrix with all ones (unweighted), \cref{eq1} degenerates into the classical form in \cite{bartlett2002rademacher,shalev2014understanding,mohri2018foundations}. Moreover, note that the derived upper bound between the expected risk and the weighted empirical risk explicitly depends on the weighted matrix $\mathbf{P}$, which implies that the weighting plays an implicit regularization role. More importantly, it also provides a theoretical guarantee that the model can still work even without any explicit regularization.

Furthermore, let $\mathcal{F}$ be a linear function class and assume that $\mathcal{L}$ is Lipschitz continuous, then we can derive the following bound.

\textbf{Theorem 2.} Let $\mathcal{F}$ be a linear function class with a bounded linear transformation $\mathbf{W}$, defined as $\mathcal{F}=\{ \mathbf{x} \mapsto \mathbf{Wx} : \left \| \mathbf{W} \right \|_F\le B_0  \}$, where $\left \| \bullet  \right \| _F$ is the Frobenius norm.  Assume $\mathcal{L} = {\textstyle \sum_{j=1}^{C}\ell(f(\mathbf{x})_j,d_\mathbf{x}^{y_j})}$, where $\ell$ is a Lipschitz continuous loss function with Lipschitz constant $L$, and $\ell$ is bounded by a constant $B$. $\left \|  \mathbf{P}\right \|_\infty = \max_{ij} \mathbf{P}_{ij}$. Then, $\forall \delta > 0$, with probability at least $1 - \delta / 2$ over the draw of an \textit{i.i.d.} sample $S$ of size $N$ from the distribution $\mathcal{D}$, the following bound holds for all $f \in \mathcal{F}$:
\begin{align}
	\mathcal{R}_{\mathcal{D}}(f) \le \hat{\mathcal{R}}_{S}(f) & + \frac{2\sqrt 2 L B_0 C^\frac{1}{2}}{\sqrt N} \max \limits_i \left \| \mathbf{x}_i \right \|_2\left \|  \mathbf{P}\right \|_\infty \nonumber \\ 
	& +   3 CB \left \|  \mathbf{P}\right \|_\infty \sqrt{\frac{log\frac{2}{\delta} }{2N} }. 
\end{align}
Theorem 2 can be proven by leveraging the main results of \cite{maurer2016vector}, and a detailed proof is provided in the Appendix. Note that, the above bound can be tighter by further assuming the loss function $\mathcal{L}$ is $\ell_\infty$ continuous, interested readers can refer to \cite{foster2019vector} for details.

\subsection{Proposed WInLDL}

Following the assumptions in Theorem 2, we design a linear weighted model named WInLDL and apply the ADMM to solve it, by which an efficient algorithm is derived. 

Specifically, given a feature matrix $\mathbf{X}$ and an observed label matrix $\widetilde{\mathbf{D}}$, let ${f}(\mathbf{X}) = \mathbf{XW}$ and $\mathcal{L}$ be the $\ell_2$ loss, then we can define the following weighted function,
\begin{align}
g(\mathbf{W})  = & \frac{1}{2} \sum_{i  = 1}^{N} \sum_{j  = 1}^{C} \mathbf{P}_{ij}((\mathbf{XW})_{ij}-\tilde{\mathbf{D}}_{ij})^2 \\  
= & \frac{1}{2} \left \| \mathbf{P}^\frac{1}{2} \odot (\mathbf{XW} - \tilde{\mathbf{D}}) \right \|_F^2,
\end{align}
where $\mathbf{W} \in \mathbb{R}^ {k \times C}$ is the transformation matrix to be optimized, and $\odot$ is the Hadamard product. Since the label distribution satisfies the probability simplex constraint, then $\mathbf{XW1}_C =\mathbf{1}_N $ and $\mathbf{XW} \ge \mathbf{0}_{N \times C}$ should hold, where $\mathbf{1}_C$ and $\mathbf{1}_N$ are column vectors of size $C$ and $N$ with all ones, ``$\ge$'' here means that all elements are greater than or equal to 0. For simplicity of notation, let $\mathbf{Q} = \mathbf{P}^\frac{1}{2}$, and $ProS(\mathbf{Z}) := \{ \mathbf{Z} \in \mathbb{R}^{N \times C} | \mathbf{Z1}_C =\mathbf{1}_N, \mathbf{Z} \ge \mathbf{0}_{N \times C} \}$. By incorporating the probability simplex constraint, the final objective function of WInLDL can be written as:
\begin{gather} 
g(\mathbf{W}) = \frac{1}{2} \left \| \mathbf{Q} \odot (\mathbf{XW} - \tilde{\mathbf{D}}) \right \|_F^2, \\
\textit{s.t.} \quad \mathbf{XW} \in ProS(\mathbf{XW}). \label{eq6}
\end{gather} 
\textbf{Remark 1.} \cref{eq6} is not an additional regularization that we impose on our model, but a constraint inherent in label distribution learning by its definition, and all label distribution learning algorithms must satisfy such a probability simplex constraint. To deal with it, we can utilize off-the-shelf projection methods\cite{wang2013projection, condat2016fast} to avoid introducing extra model hyper-parameters.

With WInLDL in hand, our main concern in the following is how to design the weighting matrix $\mathbf{Q}$. Motivated by the intuition mentioned in the Introduction, to learn an accurate label distribution for InLDL, it is crucial not to ignore the small observed degrees but to impose large weights on them, while gradually increasing the weights of the missing degrees. In order to directly exploit the degree prior of the label distribution, we subtract the observed degrees from 1 to give large weights on small degrees, formally, the weighting matrix $\mathbf{Q}$ composed of $\mathbf{Q}_{\Omega}$ and $\mathbf{Q}_{U}$ is defined as:
\begin{align}
{\mathbf{Q}}_{ij} =
\begin{cases}
1 - \widetilde{\mathbf{D}}_{ij}  & \text{if  } (i,j) \in \Omega\\  
1- \mathbf{D}_{U_{ij}}  & \text{if  } (i,j) \in U,
\end{cases}
\end{align}
where $\mathbf{D}_{U_{ij}} = \frac{1}{N} \sum_{i=1}^{N} \widetilde{\mathbf{D}}_{ij}$, that is, the missing label degrees are estimated by the mean value of the observed degrees in the corresponding column.

Moreover, to gradually increase the weights of the missing degrees, we take a number greater than 1 that increases monotonically with the number of iterations as the base of the power function. Since the observed degrees are more reliable than the missing ones, the base of its power function is set at 2, which is larger than the number ``$a$'' during the whole iterations.
\begin{align}
{\mathbf{Q}}_{ij} =
\begin{cases}
2^{(1 - \widetilde{\mathbf{D}}_{ij})}  & \text{if  } (i,j) \in \Omega \\  
a^{(1- \mathbf{D}_{U_{ij}})}  & \text{if  } (i,j) \in U, a = 1+\frac{iter}{maxIter}, \label{eqdu}
\end{cases}
\end{align} 
where $maxIter$ is the maximum iterations, in this paper, fixed at $50$. By such a design, three benefits can be obtained, 1) smaller degrees are imposed with larger weights, 2) the weights of the observed degrees are larger than the missing ones, 3) the weights of the missing degrees are gradually increased. Note that the above three benefits can also be regarded as three principles for designing the weighting matrix. Any matrix that satisfies three principles can be utilized as a weighting matrix. In the experimental section, we report the performance of various weighting matrices in Table 5. It is important to highlight that the main focus of this paper is to leverage the useful prior knowledge of label distribution to create an efficient and effective model, rather than extensively exploring the design of an optimal weighting matrix.

\subsection{Optimization}
In this subsection, we apply ADMM to design an efficient algorithm for solving the WInLDL model. Let $\mathbf{Z}=\mathbf{XW}$, the augmented Lagrangian function can be written as:
\begin{align}
\Phi = \frac{1}{2} \left \| \mathbf{Q} \odot (\mathbf{Z} - \tilde{\mathbf{D}}) \right \|_F^2 & + tr(\mathbf{\Lambda}^\top(\mathbf{XW} - \mathbf{Z})) \nonumber \\
 & + \frac{\mu}{2}\left \| \mathbf{XW} - \mathbf{Z} \right \|_F^2 ), \\
\textit{s.t.} \quad \mathbf{Z} \in ProS(\mathbf{Z}). \label{eq8}
\end{align}
where $tr$ is the trace operator, and $\mu$ is a penalty factor. Note that, $\mu$ is NOT a model hyper-parameter BUT a parameter of the ADMM algorithm, it is introduced for the convenience of optimization. In this paper, $\mu$ is fixed at 2, which does not need to be tuned, and we also conduct experiments to verify that $\mu$ only affects the convergence rate.

\textbf{Sub-problem of $\mathbf{W}$.} With $\mathbf{Z}$ and $\mathbf{\Lambda}$ fixed, $\mathbf{W}$ can be updated by,
\begin{gather}
\mathbf{W} = (\mathbf{X}^\top\mathbf{X})^{-1}(\mathbf{X}^\top(\mathbf{Z}-\frac{\mathbf{\Lambda}}{\mu} )). \label{eq9}
\end{gather}

\textbf{Sub-problem of $\mathbf{Z}$.} With $\mathbf{W}$ and $\mathbf{\Lambda}$ fixed, $\mathbf{Z}$ can be updated by,
\begin{gather}
\mathbf{Z} = \frac{\mu\mathbf{XW}+\mathbf{\Lambda}+\mathbf{Q} \odot \mathbf{Q} \odot \tilde{\mathbf{D}}}{\mathbf{Q} \odot \mathbf{Q} + \mu\mathbf{I}_{N \times C}}, \label{eq10} \\ 
\mathbf{Z} = proj(\mathbf{Z}), \label{eq11}
\end{gather}
where $\mathbf{I}_{N \times C}$ is a matrix with all ones, and the division in \cref{eq10} is element wise. \cref{eq11} projects $\mathbf{Z}$ onto the probability simplex to satisfy the constraint of \cref{eq8}, and the $proj$ is a projection operator that can be found in \cite{wang2013projection,condat2016fast}.


\textbf{Sub-problem of $\mathbf{\Lambda}$.} With $\mathbf{W}$ and $\mathbf{Z}$ fixed, $\mathbf{\Lambda}$ can be updated by,
\begin{gather}
\mathbf{\Lambda} \longleftarrow \mathbf{\Lambda} + \mu(\mathbf{XW} - \mathbf{Z}). \label{eq12}
\end{gather}

\begin{figure}[H]
	\centering
	\includegraphics[width=0.95\linewidth, height = 4cm]{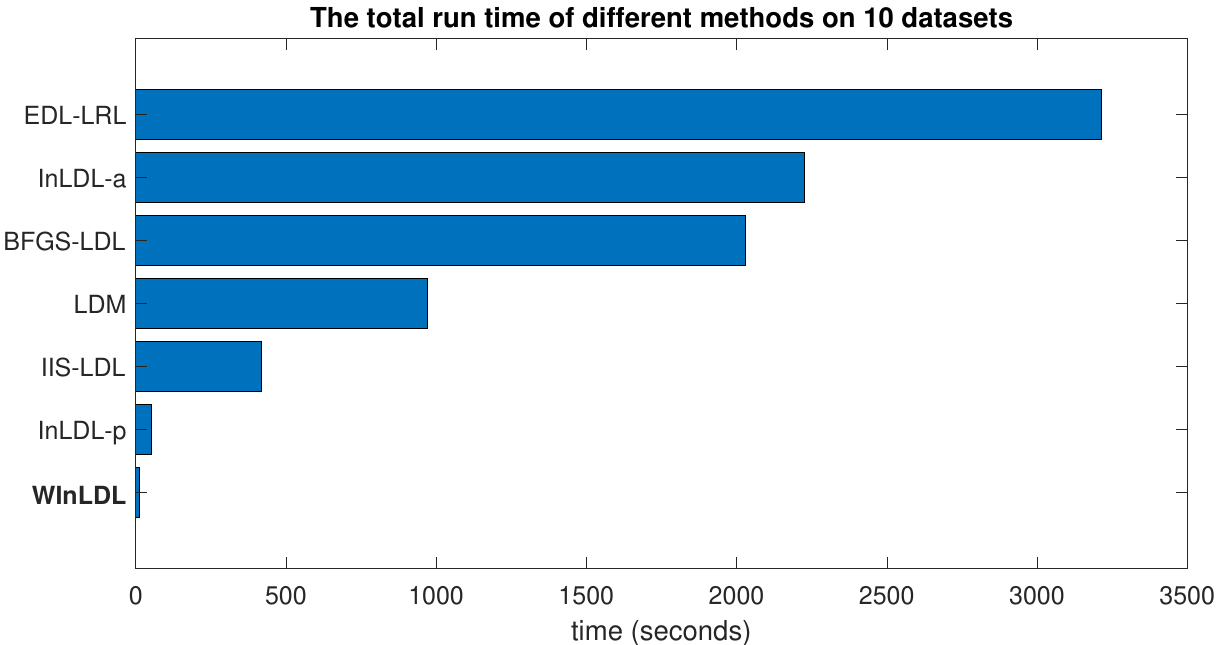} 
	\caption{The total runtime of different methods on ten real datasets, where WInLDL is our method.}
	\label{fig2}
\end{figure}

\textbf{Complexity Analysis.} The computational complexity of the ADMM algorithm is dominated by matrix multiplication and inverse operations. In each iteration, the complexity of updating $\mathbf{W}$ in \cref{eq9} is $\mathcal{O} (Nk^2)+\mathcal{O} (k^3)+\mathcal{O}(NkC)$, the complexities of updating $\mathbf{Z}$ in \cref{eq10} and \cref{eq11} are $\mathcal{O}(NkC) + \mathcal{O}(NC)$ and $\mathcal{O}(NC)$ \cite{condat2016fast}, respectively, the complexity of updating $\mathbf{\Lambda}$ in \cref{eq12} is $\mathcal{O}(NkC)$, and the complexity of updating $\mathbf{Q}_U$ in \cref{eqdu} is $\mathcal{O}(|U|)$, where $|U|$ is the cardinality of set $U$, usually smaller than $NC$. Thus, the total computational complexity is $\mathcal{O}(max(Nk^2, NkC) + k^3)$, which is linear in the number of samples $N$. In \Cref{tab1}, we list computational complexities of different methods, where `$g$' of LDM is the number of the clusters. While the computational complexity of LDM is also linear in the number of samples, it involves clustering, thus in practice, its running time is much longer than our WInLDL, details can be referred to \Cref{fig2}. 

\begin{table}[th]
	\centering
	\small
	{
		\begin{tabular}{ccc}
			\toprule[1.2pt]
			Methods                                       & Computational complexity \\  \midrule[0.8pt]     
			InLDL-a(p) 	          &  $\mathcal{O} (N^2C + C^3) $     \\
			EDL-LRL    &  $\mathcal{O} (N^2C + NkC + C^3 + k^2C^2)$  \\ 
			LDM         &  $\mathcal{O} (NC^3 + NkC + gC^4)$          \\   
			\bf{WInLDL(ours})               &  $\mathcal{O}(max(Nk^2, NkC) + k^3)$     \\ \bottomrule[1.2pt]       
		\end{tabular}
	} 
	\caption{Computational complexities of different methods.} 
	\label{tab1}
\end{table}

\section{Experiments}
\subsection{Experiments settings}
\textbf{Datasets.} In this paper, we use 10 real datasets covering fields of biology, natural scene recognition, facial expression, movie-rating, and image visual sentiment. The statistics of these datasets are summarized in \Cref{tab2}. The first five datasets are collected by Geng \cite{geng2016label}, the sixth to tenth datasets are from \cite{peng2015mixed}, \cite{liang2018scut}, \cite{yang2017learning}, \cite{li2019blended}, \cite{xie2015scut}, respectively.  

\begin{table}[h]
	\centering
	\small

	{	\begin{tabular}{ccccc}
			\toprule[1.2pt]
			Datasets            & \#Samples($N$)    & \#Features($k$)    & \#Labels($C$)          \\  \midrule[0.8pt]
			Gene         & 17892             & 36                 & 68                     \\ 
			Movie               & 7755              & 1869               & 5                      \\
			Scene      & 2000              & 294                & 9                      \\
			SBU3DFE           & 2500              & 243                & 6                      \\
			SJAFFE			    & 213               & 243                & 6                      \\
			Emotion6            & 1980              & 1000               & 7                      \\
			Fbp5500			    & 5500              & 512                & 5				      \\			
			Flickr         & 11150             & 200                & 8					  \\
			RAF\_ML             & 4908              & 200                & 6				      \\
			SCUTFBP             & 1500              & 300                & 5  				      \\
			\bottomrule[1.2pt]
		\end{tabular}
	}
     \caption{Statistics of the datasets.} 
     \label{tab2}
\end{table}

\textbf{Compared methods.} We compare our WInLDL with six methods, including two baselines named BFGS-LDL \cite{geng2016label} and IIS-LDL \cite{geng2016label}, and four state-of-the-art methods named InLDL-p \cite{xu2017incomplete}, InLDL-a \cite{xu2017incomplete}, EDL-LRL \cite{jia2019facial}, and LDM \cite{wang2023label}, respectively. BFGS-LDL and IIS-LDL are two maximum entropy models optimized with the BFGS \cite{fletcher2013practical} algorithm and the IIS \cite{della1997inducing} algorithm, respectively. InLDL-p and InLDL-a are two InLDL models that optimized by the proximal gradient descend and ADMM algorithms, respectively. EDL-LRL assumes local low-rank structure on clusters of samples, and LDM exploits both the global and local label correlations. The codes of these compared methods are shared by the original authors, and we use the best parameters suggested in their papers. 

\begin{table*}[t]
	\centering
	\small
	{
		\begin{tabular}{ccccccccc}
			\toprule[1.2pt]
			Datasets       & WInLDL                & LDM                & EDL-LRL            & InLDL-p            & InLDL-a                & BFGS-LDL            &IIS-LDL      \\ \midrule[0.8pt]
			Gene    & \textbf{0.8356(.0044)} & \underline{0.8355(.0038)} & 0.8350(.0040)       & 0.8352(.0043)       & 0.8353(.0043)           & 0.8331(.0044)  &0.8338(.0044) \\%
			Movie          & \textbf{0.9351(.0013)} & \underline{0.9329(.0017)} & 0.8517(.0125)       & 0.8824(.0019)       & 0.8886(.0017)           & 0.8475(.0016)  &0.8536(.0056) \\%
			Scene & \textbf{0.7418(.0085)} & \underline{0.7291(.0040)} & 0.6577(.0062)       & 0.6955(.0052)       & 0.6956(.0069)           & 0.6320(.0065)  &0.6613(.0064)  \\
			SBU3DFE      & \textbf{0.9417(.0021)} & {0.9224(.0018)} & 0.9191(.0022)       & 0.9329(.0035)       & \underline{ 0.9335(.0036)}     & 0.9170(.0020)  &0.9189(.0020)  \\
			SJAFFE         & \textbf{0.9517(.0043)} & 0.9341(.0018)       & 0.9343(.0019)       & \underline{ 0.9344(.0020)} & 0.9037(.0102)           & 0.9340(.0019)  &0.9309(.0040)  \\
			Emotion6       & \textbf{0.7961(.0045)} & \underline{ 0.7094(.0087)} & 0.4232(.0184)       & 0.6119(.0098)       & 0.6090(.0085)           & 0.4103(.0111)  &0.5029(.0121)   \\
			Fbp5500	       & 0.9445(.0013)          & 0.9469(.0015)       & 0.7402(.0433)       & \underline{0.9482(.0019)}  & \textbf {0.9485(.0018)} & 0.7850(.0047)  &0.8869(.0028)   \\
			Flickr   & \textbf{0.8315(.0029)} & 0.8097(.0020)       & 0.7486(.0114)       & 0.8304(.0024)       & \underline{ 0.8307(.0024)}     & 0.7537(.0089)  &0.7431(.0034) \\
			RAF\_ML        & \underline{0.8831(.0016)}     & 0.8720(.0036)       & \textbf {0.9153(.0048)} & 0.8733(.0021)   & 0.8828(.0013)           & 0.7807(.0055)  &0.6867(.0061) \\
			SCUTFBP        & \textbf{0.8138(.0054)} & \underline{0.6575(.0085)}  & 0.6435(.0060)       & 0.5663(.0104)       & 0.5909(.0087)           & 0.6447(.0078)  &0.6531(.0053) \\%
			\bottomrule[1.2pt]
		\end{tabular}
	} 
	\caption {\textit{Cosine} (the higher the better) results for incomplete setting at $50\%$ missing rate. Values in parentheses are standard deviations. The best result is in bold and the second best result is underlined.}
	\label{tab4}
\end{table*}

\begin{table*}[t]
	\centering
	\small
	{
		\begin{tabular}{ccccccccc}
			\toprule[1.2pt]
			Datasets       & WInLDL          & LDM            & EDL-LRL        & InLDL-p        & InLDL-a         & BFGS-LDL                 & IIS-LDL          \\ \midrule[0.8pt]        
			Gene    & \textbf{2.1029(.0289)} & \underline{ 2.1034(.0250)}    & 2.1095(.0264)       & 2.1091(.0284) & 2.1081(.0284) & 2.1275(.0296) & 2.1203(.0295)       \\
			Movie          & \textbf{0.5226(.0045)} & \underline{ 0.5259(.0063)}    & 0.7244(.0194)       & 0.7671(.0114) & 0.7540(.0121) & 0.7675(.0039) & 0.8013(.0240)       \\
			Scene & \underline{ 2.4634(.0125)}    & \textbf{2.4553(.0098)} & 2.4968(.0095)       & 2.5091(.0113) & 2.5137(.0099) & 2.5146(.0107) & 2.4865(.0109)       \\
			SBU3DFE      & \textbf{0.3787(.0082)} & \underline{ 0.4082(.0057)}    & 0.4169(.0062)       & 0.4373(.0130) & 0.4349(.0137) & 0.4164(.0074) & 0.4182(.0058)       \\
			SJAFFE         & \textbf{0.4096(.0187)} & 0.4160(.0094)          & \underline{ 0.4157(.0107)} & 0.4167(.0105) & 0.6375(.0247) & 0.4186(.0123) & 0.4269(.0219)       \\
			Emotion6       & \textbf{1.6148(.0163)} & \underline{1.7566(.0163)}     & 2.3937(.0143)       & 2.0360(.0092) & 2.0674(.0097) & 2.0601(.0164) & 1.9548(.0199)       \\
			Fbp5500	       & \textbf{1.2659(0.077)} & \underline{ 1.2924(.0078)}    & 1.7090(.1117)       & 1.3352(.0086) & 1.3405(.0085) & 1.4411(.0071) & 1.3404(.0060)       \\
			Flickr    & \textbf{2.1574(.0051)} & 2.1857(.0058)          & 2.1721(.0067)       & 2.2100(.0054) & 2.2096(.0055) & 2.1806(.0041) & \underline{ 2.1706(.0063)} \\
			RAF\_ML        & \textbf{1.4169(.0147)} & 1.4525(.0140)          & \underline{ 1.4468(.0129)} & 1.5252(.0111) & 1.5255(.0106) & 1.5170(.0137) & 1.5802(.0109)       \\
			SCUTFBP        & \textbf{1.4153(.0191)} & \underline{ 1.5069(.0170)}    & 1.5155(.0143)       & 1.6387(.0238) & 1.6365(.0153) & 1.5093(.0156) & 1.5071(.0141)       \\					
			\bottomrule[1.2pt]
		\end{tabular}
	}
	\caption{\textit{Clark} (the lower the better) results for incomplete setting at $50\%$ missing rate. Values in parentheses are standard deviations. The best result is in bold and the second best result is underlined.}
	\label{tab5}
\end{table*}

\textbf{Evaluation metrics.} Five commonly used metrics are applied to evaluate the performance in this paper, including \textit{Cosine}, \textit{Intersection}, \textit{Chebyshev}, \textit{Clark}, and \textit{Canberra}. The first two measure the similarity between two vectors, thus they are the higher the better, whereas the last three measure the distance between two vectors, thus they are the lower the better. For two vectors $\mathbf{p,q} \in \mathbb{R}^C$, the definitions of the five metrics are listed in the following. 1) \textit{Cosine} $\uparrow$: ${\boldsymbol{p}^{\top} \boldsymbol{q}} / ({\|\boldsymbol{p}\|_{2}\|\boldsymbol{q}\|_{2}}) $; 2) \textit{Intersection} $\uparrow$: $\sum_{i} \min \left(p_{i}, q_{i}\right)$; 3) \textit{Chebyshev} $\downarrow$: $\max _{i}\left|p_{i}-q_{i}\right|$; 4) \textit{Clark} $\downarrow$: $\sqrt{\sum_{i}\left(p_{i}-q_{i}\right)^{2} /\left(p_{i}+q_{i}\right)^{2}}$; 5) \textit{Canberra} $\downarrow$: $\sum_{i}\left|p_{i}-q_{i}\right| / (p_{i}+q_{i})$, where $\uparrow$ means the higher the better, and $\downarrow$ means the lower the better. Here we omit the KL-divergence metric just as in \cite{xu2017incomplete}, since KL-divergence is calculated by $log(d_\mathbf{x}^y / \hat{d}_\mathbf{x}^y)$, and in InLDL, the $\hat{d}_\mathbf{x}^y$ may be zero, which will make the KL-divergence meaningless.

\textbf{Incomplete settings.} Following the setting of the previous InLDL work \cite{xu2017incomplete}, we also make all elements in the label distribution matrix uniformly random missing. For fairness, we avoid adopting the setting that small degrees are more likely to be missing, as such a setting is unfair to other methods. The missing rates are vary from $10\%$ to $90\%$ with a step of $20\%$. Each method is run for five times with five random data partitions, and for each partitions, $80\%$ of the data are used for training, and the remaining $20\%$ are used for testing. For fair comparisons, in each trial, all methods are run with exactly the same missing and partitioned dataset. Finally, both the mean and the standard deviation of the results are reported.

\subsection{Main results}
In the following, we report the results of different methods at $50\%$ missing rate. Due to the page limitations, here we only list the results of the \textit{Cosine} $\uparrow$ and \textit{Clark} $\downarrow$ metrics, and results of other metrics can be found in the Appendix. 

In \Cref{tab4} and \Cref{tab5}, we show the results of the \textit{Cosine} (the higher the better) and the \textit{Clark} (the lower the better) metrics for the incomplete setting at $50\%$ missing rate on ten real datasets, respectively. For the \textit{Cosine} metric, our WInLDL ranks first on eight datasets and second on one dataset, and the average rank is $1.4$, overall, we win 57 times out of 60 comparisons, with a $95\%$ rate to win. For the \textit{Clark} metric, our WInLDL ranks first on nine datasets and second on one dataset, and the average rank is $1.1$, overall, we win 59 times out of 60 comparisons, with a $98.33\%$ rate to win. Besides, we also conduct the Nemenyi test \cite{nemenyi1963distribution,demvsar2006statistical} as the significance test, due to the page limitations, details can be found in the Appendix. From these two tables, we can conclude that WInLDL achieves better performance in most cases, which verifies its effectiveness in addressing the InLDL problem. The reason may be attributed to the weighting scheme adopted by WInLDL, which can better solve the issue that small degrees are easily overlooked, whereas other methods either only focus on mining the correlations between labels or merely adopt an entropy maximization strategy to learn the label distribution, neither of them can well address the above issue, thus leading to the suboptimal performance.

\begin{table*} [t]
	\centering
	\small
	{
		\begin{tabular}{ccccccc}
			\toprule[1.2pt]
			Datasets       & WInLDL                 & InLDL-U                  & InLDL-I            & InLDL-II                  & InLDL-Rand                     \\ \midrule[0.8pt]
			Gene    & \textbf{0.8356(.0044)} & {0.8350(.0044)}          & 0.7894 (.0055)     & \underline{ 0.8351(.0044)}       & 0.8349(.0044)            \\%
			Movie          & \textbf{0.9351(.0013)} & \underline{ 0.9349(.0014)}	   & 0.9110	(.0078)		&0.9196	(.0008)		        &0.9161	(.0010)            \\%
			Scene & \textbf{0.7418(.0085)} & \underline{ 0.7343(.0059)}      & 0.5548	(.0113)	    &0.7136	(.0065)	            &0.7065	(.0078)            \\%
			SBU3DFE      & \underline{ 0.9417(.0021)}    &\textbf{0.9426(.0017)}    & 0.9327	(.0024)	    &0.9364(.0017)              &0.9303	(.0022)            \\%
			SJAFFE         & \underline{ 0.9517(.0043)}    &\textbf{0.9555(.0043)}    & 0.9459	(.0065)	    &0.9160	(.0046)	      		&0.9086	(.0083)            \\%
			Emotion6       & \textbf{0.7961(.0045)} & \underline{0.7929	(.0048)}   & 0.7305	(.0080)	    &0.7786	(.0022)	      		&0.7619	(.0037)            \\%
			Fbp5500	       & \textbf{0.9445(.0013)} & \underline{0.9436	(.0014)}   & 0.9396	(.0008)	    &0.9192	(.0018)	      		&0.8977	(.0026)            \\%
			Flickr    & \underline{ 0.8315(.0029)}    & \textbf{0.8321(.0028)}   & 0.8107	(.0037)	    &0.8134	(.0027)	     		&0.7880	(.0030)            \\%
			RAF\_ML        & \textbf{0.8831(.0016)} & \underline{0.8798	(.0020)}   & 0.8661	(.0038)	    &0.8682	(.0025)	      		&0.8460	(.0017)            \\%
			SCUTFBP        & \textbf{0.8138(.0054)} & \underline{0.8137	(.0039)}   & 0.8009	(.0087)	    &0.7952	(.0105)	      		&0.7791	(.0108)            \\%
			\bottomrule[1.2pt]
		\end{tabular}
	}
	\caption{\textit{Cosine} (the higher the better) results for five different weighting schemes. Values in parentheses are standard deviations. The best result is in bold and the second best result is underlined.}
	\label{tab6}
\end{table*}

\subsection{Different weighting schemes}
To verify the effectiveness of imposing large weights on the small degrees, we conduct experiments on five different weighted schemes and list the results in \Cref{tab6}. The formal definitions are: (1) InLDL-U: $\mathbf{Q}_{ij} = 1$, if $(i,j) \in \Omega$, and $\mathbf{Q}_{ij} = 0$, if $(i,j) \in U$; (2) InLDL-I: $\mathbf{Q}_{ij} = \widetilde{\mathbf{D}}_{ij}$, if $(i,j) \in \Omega$, and $\mathbf{Q}_{ij} = \mathbf{D}_{U_{ij}}$, if $(i,j) \in U$; (3) InLDL-II: $\mathbf{Q}_{ij} = 2^{\widetilde{\mathbf{D}}_{ij}}$, if $(i,j) \in \Omega$, and $\mathbf{Q}_{ij} = a^{\mathbf{D}_{U_{ij}}}$, if $(i,j) \in U$, $a = 1+{iter}/{maxIter}$; (4) InLDL-Rand: $\mathbf{Q}=\mathbf{Ra}$, where $\mathbf{Ra}$ is a random matrix whose entries are uniformly distribute in $(0,1)$,  for WInLDL, referring to \cref{eqdu}. First, the results show that WInLDL consistently performs better than InLDL-I, InLDL-II, and InLDL-Rand, where InLDL-I and InLDL-II impose large weights on large degrees, and InLDL-Rand adopts random weighting. These comparisons demonstrate that imposing large weights on the small degrees is effective. Besides, WInLDL wins 7 times out of 10 comparisons with InLDL-U. Note that some LDL datasets are transformed from multi-label datasets, thus the ground truth of some degrees may be 0, and in the setting of InLDL, the missing degrees are also set to 0. In such a situation, WInLDL may put too much emphasis on the degrees whose ground truth is 0, while InLDL-U happens to treat these missing degrees as 0, which may explain why WInLDL is inferior to InLDL-U on some datasets.   

\subsection{Running time comparisons}
In this subsection, we compare the running time of the different methods and report the total time for ten datasets. All the methods are running on a Linux server with an Intel Xeon(R) W-2255 3.70GHz CPU and 64GB memory. The running time of our WInLDL is 10.95 seconds, which is orders of magnitude faster than most of the compared methods and verifies the efficiency of WInLDL.    

\subsection{Impacts study of $\mu$}
\begin{figure} [H]
	\centering
	\subfloat [Performance.] {\label{fig3a} \includegraphics[width=0.43\linewidth, height = 3cm] {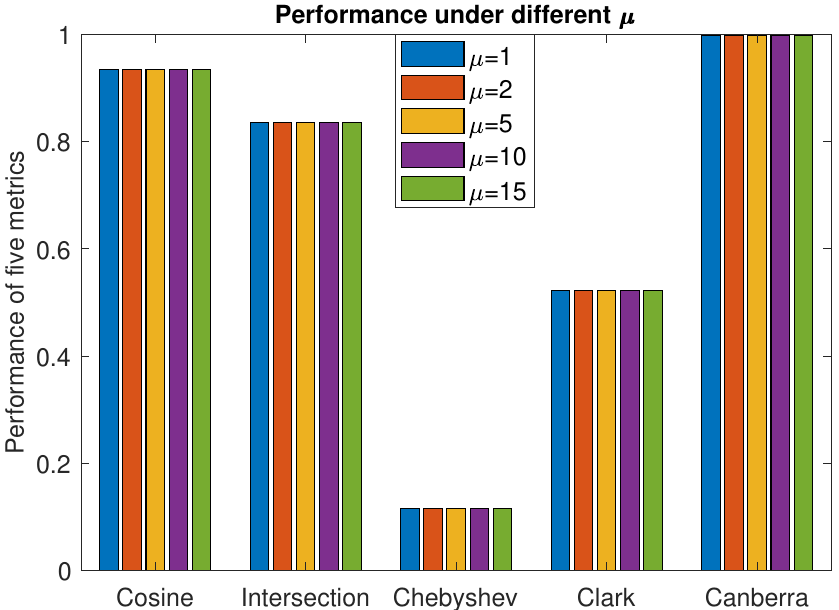}} \quad
	\subfloat [Convergence.] {\label{fig3b} \includegraphics[width=0.43\linewidth, height = 3cm] {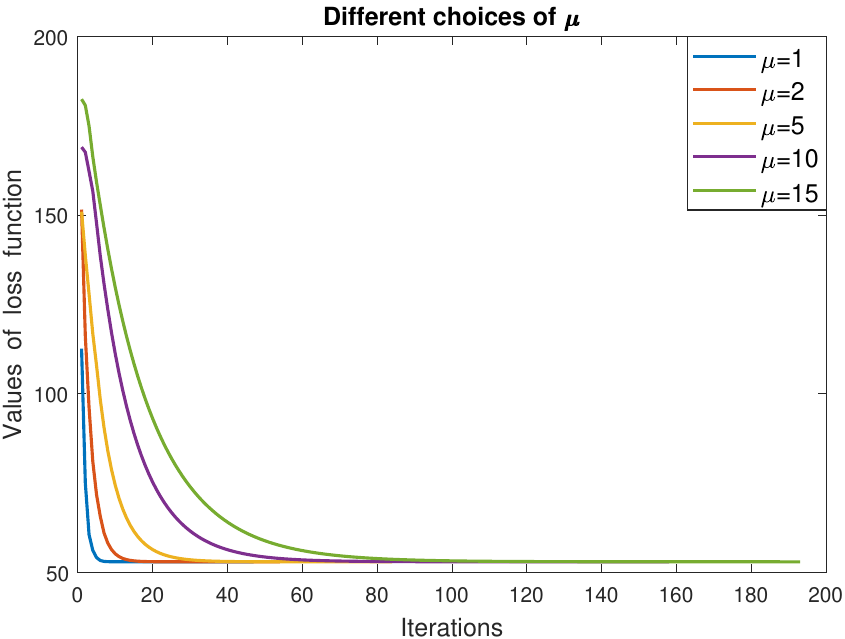}}
	\caption{The performance of all five metrics and the convergence rates under different $\mu$.} 
	\label{fig3} 
\end{figure}

In this subsection, we conduct experiments to confirm that the parameter $\mu$ in the ADMM algorithm does not change the performance of our model. From \Cref{fig3a} we find that all the five metrics remain the same regardless of variations in $\mu$, and from \Cref{fig3b} we can see that $\mu$ only affects the convergence rate. In our WInLDL model, we fix $\mu$ at 2, eliminating the need for tuning.

\section{Conclusion}


We propose an efficient, effective, and easy-to-implement model called WInLDL without any explicit regularization by properly utilizing the label distribution prior. More importantly, we present two upper bounds between the expected risk and weighted empirical risk, which explicitly depend on the $\ell_\infty$ norm of the weighted matrix. These bounds imply that such weighting plays an implicit regularization role and may explain why our model still works without any explicit regularization. By such bounds, we offer theoretical guarantees that weighting could be beneficial in LDL. Moreover, an interesting research in future work would be the theoretical guidance for how to design the optimal weighting matrix. Finally, by conducting extensive experiments, we have verified that WInLDL achieves better performance in most cases. The reason may be attributed that the weighting scheme properly utilizes the useful label distribution prior, which can well solve the issue that small degrees are easily overlooked.
\bibliography{AAAI_2024_ref}

\end{document}